\documentclass{article}

 \usepackage[preprint]{neurips_2026}

\usepackage[utf8]{inputenc}
\usepackage[T1]{fontenc}
\usepackage{hyperref}
\usepackage{url}
\usepackage{booktabs}
\usepackage{array}
\usepackage{longtable}
\usepackage{colortbl}
\usepackage{amsfonts}
\usepackage{amsmath}
\usepackage{nicefrac}
\usepackage{microtype}
\usepackage[table]{xcolor}
\usepackage{graphicx}
\usepackage[most]{tcolorbox}
\usepackage{overpic}
\usepackage{xcolor}
\usepackage{amssymb}
\usepackage{algorithm}
\usepackage{algpseudocode}
\usepackage{wasysym}
\usepackage{adjustbox}
\usepackage{subcaption}
\newcolumntype{C}{>{\centering\arraybackslash}p{1em}}

\usepackage{wrapfig}

\definecolor{darkgray}{HTML}{484B40}
\definecolor{darkred}{HTML}{712b13}
\definecolor{lightred}{HTML}{af4623}
\definecolor{darkblue}{HTML}{3c3489}
\definecolor{lightblue}{HTML}{5047b2}
\definecolor{darkgreen}{HTML}{085041}
\definecolor{lightgreen}{HTML}{265249}
\definecolor{mintcream}{HTML}{E8EBE0}
\definecolor{darkgray}{HTML}{3F4548}
\definecolor{usageDirect}{RGB}{228,244,231}
\definecolor{usageIndirect}{RGB}{255,247,223}
\definecolor{usageReserved}{RGB}{239,239,239}
\definecolor{lacFeat}{HTML}{4F46A5}   
\definecolor{lacLabel}{HTML}{006D67}  

\newcommand{\labeltag}[1]{{\fontsize{9}{8}\selectfont{\color{lacLabel}\texttt{#1}}}}
\newcommand{\feat}[1]{{\fontsize{9}{8}\selectfont{\color{lacFeat}\texttt{#1}}}}

\newcommand{\hlight}[1]{{\colorbox{lacLabel!20}{#1}}}

\renewcommand{\arraystretch}{1.08}

\newcommand{\questionbox}[1]{%
\begin{tcolorbox}[
  colback=gray!7,
  colframe=gray,
  boxrule=0.8pt,
  arc=1mm,
  left=6pt,
  right=6pt,
  top=6pt,
  bottom=6pt
]
\centering
\textbf{#1}
\end{tcolorbox}%
}

\title{Communicating Sound Through Natural Language}

\author{Emanuele Rossi \\
Sapienza University of Rome
\And
Emanuele Rodol\`a \\
Sapienza University of Rome / Paradigma
}

\begin{document}

\maketitle

\begin{abstract}
%
Natural language is widely used to describe, prompt, and control audio systems, but rarely serves as the representation carrying audio itself. We introduce \emph{lexical acoustic coding} (LAC), a framework in which pre-trained LLM sender and receiver agents transmit sound through natural language. Under fixed system prompts, the agents write their own analysis and synthesis code, communicating only through a lexical sentence, shared vocabulary, and optional symbolic music structure. The sender analyzes an input waveform into interpretable, non-learned acoustic descriptors, quantizes each with a feature-specific interval vocabulary, and verbalizes the lexical code as English. The receiver parses the sentence back into lexical-acoustic constraints and renders a waveform through closed-loop refinement. The transmitted text serves as both a rich caption and as \emph{the transport representation itself}. We frame LAC as a finite-rate lossy quantizer, exposing trade-offs between vocabulary size, rate, and fidelity. Experiments on short sounds and symbolic music transfer show that plain text preserves measurable acoustic structure while remaining interpretable, editable, and native to LLM-mediated communication.
\end{abstract}

\section{Introduction}
Natural language is usually peripheral to audio systems. It appears as metadata, supervision, prompts, or captions, while the audio itself is transported as waveforms, continuous latents, or learned codec tokens \citep{zeghidour2021_soundstream,defossez2022_encodec,borsos2022_audiolm,agostinelli2023_musiclm}. In this work, we study a different design point: a sound is projected onto an interpretable acoustic feature space, discretized into a fixed lexical code, and rendered as ordinary English prose. The sentence thus becomes the \emph{transport representation} of the sound it describes\footnote{See demo page: {\scriptsize\url{https://erodola.github.io/lac-demo/}}}.

The motivating question of this paper is whether a sufficiently structured vocabulary can carry enough acoustic information to support a \emph{usable} round trip between language-model agents.

Our starting point is that descriptor-based audio analysis already provides a compact, interpretable vocabulary for many perceptually salient aspects of sound, including spectral shape, temporal envelope, and harmonic structure \citep{lartillot2007_mirtoolbox,mcfee2015_librosa,caetano2019_audio_descriptors_timbre}. At the same time, work on timbre semantics and audio production shows that humans routinely describe sound through stable verbal descriptors, and such language can support actionable control \citep{saitis2019_semantics_timbre,cartwright2013_socialeq,roche2021_metallic,venkatesh2022_word_embeddings_eq,kumar2024_sila}. Recent speech and audio language models move in a related direction by learning text-aligned or semantically factorized audio tokenizations \citep{tseng2025_taste,wang2025_tadicodec,dang2026_tada,yang2026_uniaudio2,yu2024_salmonn_omni}. 
We ask a different question:
\questionbox{Can language be structured so that it carries measurable acoustic information?}

The setting we consider is fully agentic, and has a one-time setup phase and a per-sound transmission phase.
A sender and a receiver LLM operate under fixed system prompts, but each writes its own code for analysis and synthesis. Sender and receiver are off-the-shelf and not trained for this task. 

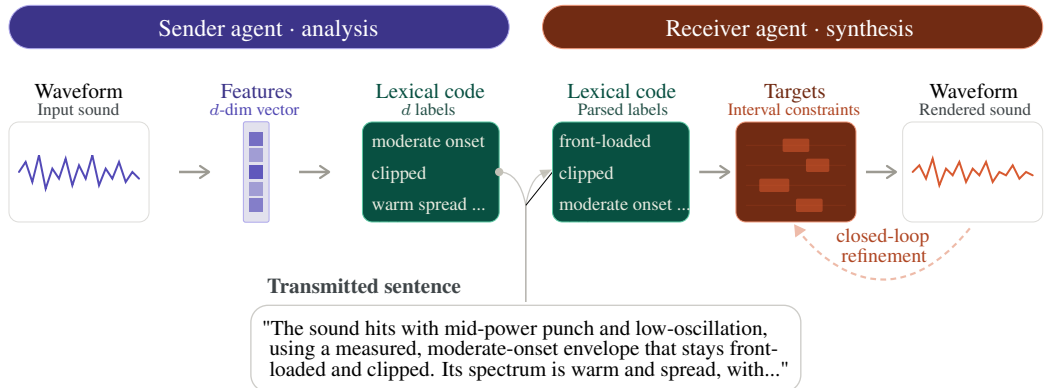
\begin{figure}[t]
    \centering
    \begin{overpic}[width=.98\linewidth]{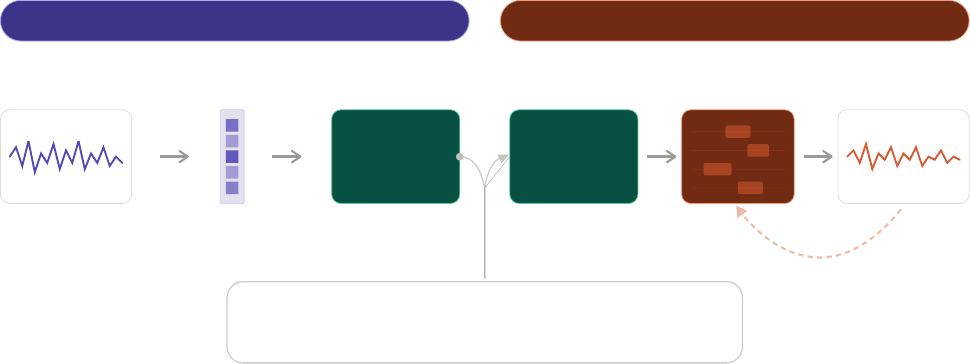}
    \put(14.5,34.5){\fontsize{9}{8}\selectfont \color{white}Sender agent $\cdot$ analysis}
    \put(63.5,34.5){\fontsize{9}{8}\selectfont \color{white}Receiver agent $\cdot$ synthesis}
    \put(35.1,23.7){\fontsize{7}{8}\selectfont \color{mintcream}moderate onset}
    \put(35.1,20.7){\fontsize{7}{8}\selectfont \color{mintcream}clipped}
    \put(35.1,17.7){\fontsize{7}{8}\selectfont \color{mintcream}warm spread ...}
    \put(53.2,23.7){\fontsize{7}{8}\selectfont \color{mintcream}front-loaded}
    \put(53.2,20.7){\fontsize{7}{8}\selectfont \color{mintcream}clipped}
    \put(53.2,17.7){\fontsize{7}{8}\selectfont \color{mintcream}moderate onset ...}
    \put(25,9.65){\fontsize{8}{8}\selectfont \color{darkgray}\textbf{Transmitted sentence}}
    \put(2.5,28.5){\fontsize{8}{8}\selectfont \color{black}Waveform}
    \put(2.8,26.8){\fontsize{6.5}{8}\selectfont \color{darkgray}Input sound}
    \put(20.5,28.5){\fontsize{8}{8}\selectfont \color{darkblue}Features}
    \put(19.4,26.8){\fontsize{6.5}{8}\selectfont \color{lightblue}$d$-dim vector}
    \put(35.5,28.5){\fontsize{8}{8}\selectfont \color{darkgreen}Lexical code}
    \put(37.9,26.8){\fontsize{6.5}{8}\selectfont \color{lightgreen}$d$ labels}
    \put(54,28.5){\fontsize{8}{8}\selectfont \color{darkgreen}Lexical code}
    \put(55,26.8){\fontsize{6.5}{8}\selectfont \color{lightgreen}Parsed labels}
    \put(73,28.5){\fontsize{8}{8}\selectfont \color{darkred}Targets}
    \put(69.5,26.8){\fontsize{6.5}{8}\selectfont \color{lightred}Interval constraints}
    \put(89,28.5){\fontsize{8}{8}\selectfont \color{black}Waveform}
    \put(88,26.8){\fontsize{6.5}{8}\selectfont \color{darkgray}Rendered sound}
    \put(80,14.6){\fontsize{7.5}{8}\selectfont \color{lightred}closed-loop}
    \put(80.5,12.6){\fontsize{7.5}{8}\selectfont \color{lightred}refinement}
    \put(24.5,5.6){\fontsize{8}{8}\selectfont "The sound hits with mid-power punch and low-oscillation,}
    \put(25.2,3.6){\fontsize{8}{8}\selectfont using a measured, moderate-onset envelope that stays front-}
    \put(25.2,1.6){\fontsize{8}{8}\selectfont loaded and clipped. Its spectrum is warm and spread, with..."}
    \end{overpic}
    \caption{\textbf{LAC pipeline.} A waveform is analyzed into a short descriptor, quantized into a lexical code, and verbalized as an English sentence; the sentence then crosses the channel. The receiver parses it back into labels, inverts each label to an interval target, and renders a waveform via a decoder with closed-loop refinement. Not a single binary data byte is ever transmitted end-to-end; complete examples of sounds and transmitted sentences are available in the demo page.\label{fig:pipeline}}
\end{figure}

In the \emph{setup} phase, sender and receiver are given the same vocabulary; this vocabulary can be human-authored and shared once, or generated by the source agent and transmitted once before any sounds are sent. In the \emph{per-sound} phase, the sender analyzes an input sound, maps it to a $d$-dimensional lexical code (with small $d$) using the shared vocabulary, verbalizes that code as an English sentence, and sends only that sentence. The receiver maps the sentence back to the same $d$-label code and renders an approximate waveform from the decoded acoustic constraints; see Figure \ref{fig:pipeline}.

No binary audio payload, learned latent, or non-ASCII side channel is transmitted during either phase. The payload is a human-readable sentence; its reversibility comes from the shared vocabulary and from the constraint that the sentence preserves each lexical term unambiguously.

\paragraph{Scope and objective.}
Once a sound is projected into lexical acoustic coordinates, the goal is no longer exact sample recovery. Instead, the decoder operates {generatively}: it produces a waveform consistent with the transmitted lexical description, which can be modified and updated to generate new sounds if desired. In this sense, LAC is closer to a semantic communication system than to a conventional codec \citep{jiang2023_semantic_foundation_models}. The representation deliberately trades bit-level invertibility for properties that ordinary codecs do not jointly prioritize: human readability, acoustic interpretability, text-native transport, and compatibility with agentic code generation at both ends of the channel.

This places LAC between captioning and compression. Captions are readable but too weak for reconstruction; codecs are invertible but opaque. Our aim is a third object: a \emph{human-readable acoustic code} whose sentence form remains informative about how the sound should \emph{sound}, while still enabling reconstruction through an explicit inverse mapping.

We theoretically view LAC as a finite-rate lossy quantizer, exposing the trade-off between vocabulary size, rate, and reconstruction fidelity. As an application, we consider music transfer where structure is transmitted separately (e.g. in ABC notation or free-form language) while timbre is carried by LAC.

\paragraph{Contribution.}
This paper makes three contributions.
\begin{itemize}
    \item We formalize \emph{lexical acoustic coding} (LAC): a framework quantizing acoustic descriptors into a short lexical code, verbalized for plain-text English transmission between LLM agents, and we give basic finite-rate and lossy-quantizer results for the resulting channel.
    \item We introduce an agentic setting where sender and receiver write their own analysis and synthesis code under fixed prompts, while communicating only through the lexical sentence, vocabulary, and optional symbolic music structure.
    \item We introduce hybrid reconstruction with interval-aware closed-loop refinement, and show that LAC supports both short isolated sounds and symbolic music transfer, where song structure is preserved externally and timbre passes through the lexical channel.
\end{itemize}

The system occupies a distinct point in the design space of audio representations: \emph{not} a replacement for neural codecs on raw rate--distortion grounds, but an interpretable, inspectable, plain-text acoustic representation that humans can read and preserve, language models can manipulate, and decoders can render into sound. 

\section{Related Work}
\paragraph{Descriptor-based and perceptual audio representations.}
Our method builds on a long tradition of interpretable audio analysis using handcrafted descriptors. In speech and MIR,  spectral statistics, envelope features, harmonicity measures, and cepstral summaries have long provided compact proxies for waveform structure \citep{davis1980_mfcc,lartillot2007_mirtoolbox,mcfee2015_librosa}. These representations remain useful when editability, interpretability, and direct links to known acoustical quantities are preferred over end-to-end latent variables. In parallel, perceptual studies examine how listeners describe sound with adjectives and metaphors, and how such language relates to measurable acoustic structure \citep{saitis2019_semantics_timbre,cartwright2013_socialeq,roche2021_metallic}. Recent work has also explored mapping text to concrete audio parameters \citep{venkatesh2022_word_embeddings_eq}, including equalization and related sound attributes \citep{kumar2024_sila}. 

Our work is grounded in both traditions: it starts from standard acoustic descriptors, but then lexicalizes them into prose that is intended to remain meaningful to a human or artificial agent.

\paragraph{Language as control vs language as transport.}
Several prior systems use language to \emph{control} sound, rather than to \emph{carry} sound. SocialEQ crowdsourced actionable equalization (EQ) descriptors from users, explicitly seeking mappings between terms such as ``warm'' and EQ operations \citep{cartwright2013_socialeq}. Subsequent work used word embeddings to predict EQ settings from semantic descriptors, including descriptors not seen during training \citep{venkatesh2022_word_embeddings_eq}. In sound synthesis, perceptually grounded latent spaces have been designed to support verbal control over timbral attributes such as metallic, warm, breathy, or percussive \citep{roche2021_metallic}. More recently, SILA augments text-to-audio generation with explicit control over acoustic characteristics such as loudness, pitch, reverb, brightness, noise, and duration \citep{kumar2024_sila}. 

These approaches establish that natural language can be an effective \emph{interface} for audio manipulation and generation. Our setting is different in that the sentence is neither a prompt, a pure caption, or user control signal. Rather, the sentence is the transmitted representation itself: a lossy, human-readable code from which a receiver reconstructs a quantized acoustic feature vector and then audio.

\paragraph{Learned audio codecs and audio-language tokenization.}
A different line of work learns compact representations for audio. Neural codecs such as SoundStream \citep{zeghidour2021_soundstream} and EnCodec \citep{defossez2022_encodec} compress waveforms into learned discrete codes optimized jointly with neural decoders. Language-audio generative models such as AudioLM \citep{borsos2022_audiolm} and MusicLM \citep{agostinelli2023_musiclm} then model sequences of such learned audio tokens to enable long-range generation and text conditioning. Recent speech-language models have pushed this further toward text-aligned or semantically factorized tokenizations. TASTE  \citep{tseng2025_taste} learns a text-aligned speech tokenizer through attention-based aggregation and a reconstruction objective; TaDiCodec \citep{wang2025_tadicodec} uses a text-aware diffusion codec to achieve low-rate speech tokenization; TADA \citep{dang2026_tada} proposes one-to-one synchronization between text tokens and acoustic features; UniAudio~2.0 \citep{yang2026_uniaudio2} factorizes audio into reasoning and reconstruction tokens; and SALMONN-omni \citep{yu2024_salmonn_omni} removes explicit codec injection in a full-duplex speech LLM. 

These works are conceptually related because they seek compact language--audio representations. However, they rely on \emph{learned} latents, internal tokenizers, or end-to-end neural decoders. In contrast, LAC uses an explicit lexical code over classical acoustic descriptors, transmitted in ordinary English and inverted through deterministic analysis/synthesis.

\paragraph{Symbolic structure and explicit analysis/synthesis pipelines.}
Our formulation is also related to work that separates symbolic musical structure or control information from waveform rendering. JAMS \citep{humphrey2014_jams}, for example, provides structured, machine-readable annotations for music and audio research, while factorized music and audio pipelines separate event structure, control trajectories, and synthesis, including DDSP-style differentiable control \citep{engel2020_ddsp,wudsp2022_midi_ddsp} and MIDI-conditioned performance modeling \citep{hawthorne2019_maestro}. 

These lines of work support the broader idea that audio need not be represented only as waveform samples, but can also be mediated by an intermediate representation. In our setting, that intermediate representation is natural language. More specifically, the proposed framework can transmit both an acoustic description of the sound and, when available, an explicit symbolic representation of the music, such as MIDI-like structure. The key difference is therefore not the use of an intermediate representation as such, but the use of \emph{natural language} as the transport layer for acoustic content.

\paragraph{Natural language as a communication channel.}
More broadly, our work connects to the emerging literature on semantic communication with foundation models, which studies how shared priors and world knowledge can shift communication away from raw bits toward higher-level representations \citep{jiang2023_semantic_foundation_models}. It is also adjacent to recent work on linguistic steganography and covert channels, where natural language is used to carry arbitrary payloads robustly under paraphrasing or distributional constraints \citep{gaure2024_covert_channels,perry2025_robust_steganography,norelli2026llms}. 

Yet our goal is fundamentally different from covert transmission. We do not seek to hide arbitrary bits in fluent cover text. Instead, we seek an \emph{overt}, human-readable, perceptually grounded channel in which the transmitted sentence remains informative about how the sound actually \emph{sounds}.

\paragraph{Our positioning.}
We sit at the intersection of descriptor-based timbre analysis, language-based sound control, and learned audio tokenization, but differ from each (see Table \ref{tab:lac_comparison}). We lexicalize standard acoustic descriptors into a constrained natural-language code, use prose as the representation itself rather than merely a control interface, and replace learned latent tokens with a human-interpretable ASCII transport layer. To our knowledge, prior work has not combined these properties in a single representation that is descriptor-grounded, intelligible to experts, and invertible for lossy waveform reconstruction through LLM-mediated sender/receiver pipelines.

\begin{table}[t]
\centering
\small
\renewcommand{\arraystretch}{0.95}
\newcommand{\yes}{\textcolor{teal}{\CIRCLE}}
\newcommand{\no}{\textcolor{lightgray}{\Circle}}
\newcommand{\half}{\textcolor{teal}{\LEFTcircle}}
\newcommand{\rot}[1]{\rotatebox{60}{\small #1}}
\setlength{\tabcolsep}{7pt}
\begin{adjustbox}{max width=\textwidth}
\begin{tabular}{lCCCCCCCC}
\textbf{Method}
  & \rot{\fontsize{7}{8}\selectfont \textbf{Human readability}}
  & \rot{\fontsize{7}{8}\selectfont \textbf{LLM-native transport}}
  & \rot{\fontsize{7}{8}\selectfont \textbf{Semantic editing}}
  & \rot{\fontsize{7}{8}\selectfont \textbf{Acoustic interpretability}}
  & \rot{\fontsize{7}{8}\selectfont \textbf{Training-free}}
  & \rot{\fontsize{7}{8}\selectfont \textbf{Generative decoding}}
  & \rot{\fontsize{7}{8}\selectfont \textbf{Bandwidth efficiency}}
  & \rot{\fontsize{7}{8}\selectfont \textbf{Bit-wise reconstruction}} \\
\midrule
Lossless codec (FLAC, WAV)                  & \no   & \no   & \no   & \no   & \yes  & \no   & \no   & \yes \\
Handcrafted descriptors (MFCC, spectral centroid) & \no   & \no   & \no   & \yes  & \yes  & \no   & \yes  & \no  \\
Neural codec (EnCodec, SoundStream)         & \no   & \half   & \no   & \no   & \no   & \yes  & \yes  & \no  \\
Audio-language tokenizers (AudioLM, TASTE)  & \no   & \half & \half & \no   & \no   & \yes  & \yes  & \no  \\
Unconstrained text caption                  & \yes  & \yes  & \yes  & \half & \yes  & \yes  & \no   & \no  \\
\midrule
\textbf{LAC (this paper)}                   & \yes  & \yes  & \yes  & \yes  & \yes  & \yes  & \half & \no  \\
\bottomrule
\end{tabular}
\end{adjustbox}
\caption{Comparison of audio representation methods.
\yes\,=\,yes; \half\,=\,partial; \no\,=\,no. Axis definitions and per-method justifications in Appendix \ref{app:comparison-table}.}
\label{tab:lac_comparison}
\end{table}
\section{Method}
Our method follows a communication protocol with shared vocabulary: each input sound is analyzed and mapped into the vocabulary, verbalized as a sentence, transmitted to a decoder, mapped back to numerical values, and re-synthesized as an audio waveform. Encoder and decoder are pre-trained language models, and the transmission happens through a pure text channel.

\subsection{Shared vocabulary}
\label{sec:method-vocabulary}
The first step in the transmission is the analysis of the input sound into numerical features. The feature choice is done once at the beginning, and encoded into a feature set $\mathcal{F}$; for example:
\begin{center}
$\mathcal{F}=\{$ \feat{rms\_energy}\,, \feat{decay\_time}\,, \feat{spectral\_centroid}\,, $\dots$ $\}\,.$
\end{center}

Let $d=|\mathcal{F}|$ be the number of features. For each feature $f_i$, we define $\mathcal{A}_i$ to be its corresponding lexical alphabet; the full lexical state space is then:
\begin{align}
    \mathcal{L} := \mathcal{A}_1 \times \cdots \times \mathcal{A}_d\,.
\end{align}
Each individual alphabet $\mathcal{A}_i$ has a different size depending on the feature, and is fixed a priori. 
For example, the \feat{rms\_energy} feature has the following alphabet of size 5:

\begin{center}
$\mathcal{A}_{\feat{rms\_energy}}=\{$ \labeltag{whisper}\,, \labeltag{hushed}\,, \labeltag{mid-power}\,, \labeltag{forceful}\,, \labeltag{thunderous} $\}\,.$
\end{center}

The alphabets $\mathcal{A}_i$ admit freedom in the lexical choices, and might be human-written or generated once by the sender agent. In this paper, we use an agent-generated lexicon.

Feature set $\mathcal{F}$ and lexical state space $\mathcal{L}$ are shared at the beginning, as part of a vocabulary
\begin{equation}
    \mathcal{V} = \{f_i,\mathcal{A}_i, E_i, R_i, I_i\}_{i=1}^{d} \,,
    \label{eq:vocabulary}
\end{equation}
where $E_i:\mathbb{R}\to \mathcal{A}_i$ maps a feature value to a lexical label (Section \ref{sec:enc}), while $R_i:\mathcal{A}_i\to \mathbb{R}$ and $I_i: \mathcal{A}_i \to \mathbb{R}^2$ map the label back to the feature's interval midpoint and bounds (Section \ref{sec:dec}).

\paragraph{Remark.}
The full vocabulary $\mathcal{V}$ is transmitted as \emph{pure text}, including the chosen feature set, the lexical mapping, and agent instructions on how to implement each feature. See Appendix \ref{app:feature-provenance} and \ref{app:feature-vocab-mapping} for the specific information that we used in our prompts.

Once the vocabulary is shared, the encode--transmit--decode pipeline follows.

\subsection{Encoder}\label{sec:enc}
The encoder first converts the waveform into a fixed vector of acoustic features, and then lexicalizes each coordinate independently. We start by defining the acoustic feature extractor.

\paragraph{Acoustic features.}
Given a mono waveform $s \in \mathbb{R}^{T}$, the encoder computes a feature vector
\begin{equation}
    x = F(s) \in \mathbb{R}^{d}\,,
\end{equation}
where each coordinate corresponds to a different feature $f_i$, $i=1,\dots,d$.

The feature extractor $F(s)$ uses off-the-shelf components and common audio toolkits such as \texttt{librosa} \citep{mcfee2015_librosa} to keep the system end-to-end, training-free, and reproducible. This way, every coordinate has an acoustically meaningful interpretation that can be named and transmitted.

We compute $d=47$ features, organized as 7 temporal, 7 spectral, 7 harmonic, and 26 psychoacoustic ones.
Appendix \ref{app:feature-list} reports their short description.

\paragraph{Lexical code.}
Each coordinate is lexicalized independently:
\begin{equation}
    \ell = (\ell_1,\ldots,\ell_{d}),
    \qquad
    \ell_i = E_i(x_i).
    \label{eq:lexical-code}
\end{equation}
The map $E_i:\mathbb{R}\to\mathcal{A}_i$ is implemented as a feature-specific interval table. 

For example, an RMS value in $[0.10,0.30)$ maps to \labeltag{mid-power}. The full feature set and vocabulary mapping are reported in Appendices~\ref{app:feature-list} and~\ref{app:feature-vocab-mapping}. We treat these choices as specific instantiations of the LAC framework rather than core contributions: they were generated by an agent on a best-effort basis and are not optimized; we leave the search for better feature subsets and mappings to future work.

\subsection{Sentence transport}
The ordered code $\ell$ is not sent as a comma-separated list. It is converted into an English sentence
\begin{equation}
    q = V(\ell)
    \label{eq:sentence-verbalization}
\end{equation}
that contains all $d$ lexical terms in recoverable form. 

The verbalizer $V$ may add ordinary grammatical material, but it may not delete, merge, paraphrase, or ambiguously rename any term. This is what distinguishes the transmitted sentence from a loose prose caption: it is readable English, but it remains an \emph{injective} carrier for the acoustic code.

The inverse map is a parser $U$ such that, for all $\ell\in\mathcal{L}$:
\begin{align}
    U(V(\ell)) = \ell\,.
\end{align}

\paragraph{Example.} Consider a short sequence $\ell$ with $d=3$:
\begin{center}
$\ell = ($ \labeltag{thunderous}\,, \labeltag{swift-onset}\,, \labeltag{short-decay} $)\,.$
\end{center}
This can be written in a sentence $q$:
\begin{center}"A \hlight{thunderous} sound with a \hlight{swift onset} and a \hlight{short decay}."\end{center} 
Since the sentence is written by the sender LLM, it might differ at each run in terms of prose.

The receiver applies the inverse parser
\begin{equation}
    \ell = U(q) \in \mathcal{L}
    \label{eq:sentence-parsing}
\end{equation}
to recover the $d$-slot lexical code before synthesis. Thus the per-sound payload is the sentence $q$, while the recoverable object carried by that sentence is the full lexical code $\ell$.

\paragraph{Remark (finite-rate bottleneck).}
While the sentence $q$ may be verbose, its recoverable acoustic content cannot be hidden in the prose itself, and the reconstruction quality is instead limited by the information carried by lexical state $\ell$. 

We quantify this bound via $B_{\max}$, the worst-case budget (in bits) needed to represent a lexical state:
\begin{equation}
    B_{\max}
    :=
    \log_2 |\mathcal{L}|
    =
    \sum_{i=1}^d \log_2 |\mathcal{A}_i|.
    \label{eq:finite-budget}
\end{equation}
Formally, let $S$ be a random source sound, $L$ its lexical code, and $\widetilde{S}$ any reconstruction computed from $L$ alone. Then
\[
S \to L \to \widetilde{S}
\]
is a Markov chain, so the data processing inequality gives
\begin{equation}
    I(S;\widetilde{S})
    \le
    I(S;L)
    =
    H_{\mathrm{Sh}}(L)
    \le
    \log_2 |\mathcal{L}|.
    \label{eq:info-bottleneck}
\end{equation}
If only a subset of lexical states is ever realized, this sharpens to
\[
H_{\mathrm{Sh}}(L)\le \log_2 |\operatorname{supp}(L)|.
\]
Thus $\log_2 |\mathcal{L}|$ is a distribution-free ceiling on the per-sound lexical payload, while $H_{\mathrm{Sh}}(L)$ is the average source information actually carried by that payload under a chosen source distribution.

\subsection{Decoder}\label{sec:dec}
The decoder first maps the transmitted sentence $q$ back to a finite lexical code, $\ell=U(q)$. Then, it treats the recovered labels as interval-valued acoustic constraints. The labels are converted into representative synthesis targets, then rendered as audio with a deterministic hybrid synthesizer. 

A refinement step closes the loop by re-analyzing the synthesized audio and adjusting a small set of features until the entire feature vector better satisfies the transmitted intervals.

\paragraph{Label inversion.}
Each recovered label provides two objects, i.e. a \emph{midpoint} value and interval:
\begin{equation}
    \tilde{x}_i = R_i(\ell_i)\,,
    \qquad
    [a_i,b_i) = I_i(\ell_i)\,.
    \label{eq:label-inversion}
\end{equation}
The representative midpoint $\tilde{x}_i$ is used for synthesis. For example, for the RMS label \labeltag{mid-power}, corresponding to $[0.10,0.30)$, the value $\tilde{x}_i = 0.20$ is used.

The $d$ representatives (one per feature) are bundled into decoder parameters
\begin{equation}
    \theta = B(\tilde{x}_1,\ldots,\tilde{x}_{d})\,,
    \label{eq:decoded-parameters}
\end{equation}
where $B$ groups them into temporal, spectral, harmonic, Bark-band, and psychoacoustic targets. 

A deterministic seed $\sigma = \mathrm{hash}(\ell)$ is also computed from the recovered lexical code, and later used to initialize the waveform renderer.

\paragraph{Remark (feature-space quantization).}
The lexical stage is a deterministic lossy quantizer in feature space. For any coordinate whose label $\ell_i$ corresponds to a bounded interval $[a_i,b_i)$, midpoint decoding is minimax-optimal, minimizing the worst-case absolute error within the bin:
\begin{equation}\label{eq:midpoint-minimax}
\tilde{x}_i = \tfrac{a_i+b_i}{2} = \operatorname*{arg\,min}_{r\in[a_i,b_i)} \sup_{x\in[a_i,b_i)} |x-r|, \qquad |x_i-\tilde{x}_i| \le \tfrac12(b_i-a_i).
\end{equation}
Hence, for any nonnegative weights $\alpha_i$,
\[
    \sum_{i\in\mathcal{B}} \alpha_i (x_i-\tilde{x}_i)^2 \le \tfrac14 \sum_{i\in\mathcal{B}} \alpha_i (b_i-a_i)^2,
\]
where $\mathcal{B}$ is the set of bounded coordinates. This bounds the distortion from lexicalization alone; the rendered waveform may incur additional error.

\paragraph{Waveform renderer.}
The waveform synthesizer is a hybrid renderer that combines harmonic, modal, and noise-based components. Given decoded acoustic targets $\theta$, an internal control vector $c$, and a deterministic seed $\sigma$, it produces a waveform
\begin{equation}
    \tilde{s} = G(\theta,c,\sigma)\,,
    \label{eq:hybrid-renderer}
\end{equation}
where $\sigma$ fixes the stochastic parts of the renderer so that repeated evaluations remain reproducible. The renderer is called repeatedly during refinement.

Concretely, $G$ instantiates a harmonic sine bank for pitched content, a resonant modal layer, seeded body and transient noise, attack--decay envelopes, Bark-domain equalization, broad spectral sculpting, and final RMS normalization.

The control vector $c$ contains renderer-internal steering variables (e.g. envelope scaling, modal density, spectral shaping) that control the synthesizer in low dimension. Many decoded features in $\theta$ are rendered directly, while others are matched only indirectly through the coupled effect of these controls. As a result, changing one control typically moves several measured features at once.

The lexical code therefore specifies the acoustic region to be matched, while the control vector provides a compact way of steering a generic synthesizer toward that region; refinement then checks the rendered waveform against the full set of target intervals. Additional implementation details are deferred to Appendix~\ref{app:decoder-details}. 

\begin{wrapfigure}[18]{r}{0.5\textwidth}
  \vspace{-0.3cm}
  \begin{minipage}{0.50\textwidth}
    \captionsetup{type=algorithm}
    \caption{Receiver-side decoding}
    \label{alg:decode}
    \small
    \begin{algorithmic}[1]
\Require Transmitted sentence $q$, shared vocabulary
\Ensure Synthesized waveform $\tilde{s}$
\State Recover the lexical code $\ell \leftarrow U(q)$ and validate
\State Decode acoustic targets $\theta$ from $\ell$ using Eq. \eqref{eq:label-inversion}
\State Compute the deterministic seed $\sigma \leftarrow \mathrm{hash}(\ell)$
\State Initialize renderer controls $c_0$ from $\theta$
\For{each $c$ visited by the search}
    \State Render $\hat{s} \leftarrow G(\theta,c,\sigma)$ (Eq.~\eqref{eq:hybrid-renderer})
    \State Re-extract features $\tilde{x} \leftarrow F(\hat{s})$ (Eq.~\eqref{eq:reanalyze})
    \State Score the candidate (Eq.~\eqref{eq:refinement-objective})
\EndFor
\State Select $c^\star$: prefer fewer violated lexical bins, then lower score
\State $\tilde{s} \leftarrow G(\theta,c^\star,\sigma)$
\State \Return $\tilde{s}$
    \end{algorithmic}
  \end{minipage}
  \vspace{-\intextsep}
\end{wrapfigure}
\paragraph{Closed-loop refinement.}
Given decoded targets ${\theta\in\mathbb{R}^d}$, the renderer controls are initialized at $c_0$ and refined during optimization. For any candidate $c$, the decoder renders audio and re-extracts the features:
\begin{equation}
    \tilde{x} = F\!\bigl(G(\theta,c,\sigma)\bigr)\,.
    \label{eq:reanalyze}
\end{equation}
It then scores the candidate against the transmitted description:
\begin{equation}
    J(c)
    =
    \operatorname{mismatch}(\tilde{x},\ell)
    +
    \operatorname{reg}(c,c_0)\,.
    \label{eq:refinement-objective}
\end{equation}
The mismatch term is small when re-extracted features lie inside the lexical intervals implied by $\ell$, and grows outside them. The regularizer keeps controls near $c_0$. We optimize $J$ with derivative-free Powell search \citep{powell64}; Algorithm \ref{alg:decode} gives the full procedure.

The final waveform is chosen from Powell’s candidates as the one violating the fewest lexical bins, using $J(c)$ as a tie-breaker. This aligns refinement with the transmitted discrete description rather than over-optimizing a smooth proxy.

\section{Experiments}
\label{sec:experiments}
We evaluate the full LAC pipeline with GPT-5.3-Codex sender and receiver agents at \texttt{xhigh} reasoning effort, using a single-turn protocol conditioned on the shared vocabulary.
All runs used one Apple M2 MacBook Air with 8 CPU cores and 24 GB unified memory, totaling under 3 hours wall-clock.

\begin{wraptable}[9]{r}{0.26\linewidth}
\vspace{-0.45cm}
\centering
\captionsetup{font=footnotesize}
\scriptsize
\setlength{\tabcolsep}{4pt}
\renewcommand{\arraystretch}{0.95}
\caption{Dataset statistics.}
\label{tab:dataset-statistics}
\begin{tabular}{@{}lr@{}}
\toprule
Stat & Value \\
\midrule
Songs & \textbf{307} \\
Samples & \textbf{3707} \\
Mean samples / song & \textbf{12.07} \\
Min sample dur. (s) & \textbf{0.0002} \\
Mean sample dur. (s) & \textbf{0.326} \\
Max sample dur. (s) & \textbf{1.997} \\
Looped samples & \textbf{2092} \\
Mean song dur. (s) & \textbf{141.88} \\
\bottomrule
\end{tabular}
\end{wraptable}
\subsection{Dataset}
We constructed a new \emph{tracker music} dataset. Tracker songs are self-contained modules that store both a note sequence (in ABC-like notation) and the short audio samples used to play it, commonly in formats such as \texttt{.mod}, \texttt{.xm}, and \texttt{.it}. They are convenient for our setting because the samples are already isolated timbral events, while the module structure still enables song-level transfer experiments.

We collected $\sim$300 public modules and used a Python library for parsing\footnote{Archive: {\scriptsize\url{https://amp.dascene.net/}}; \hspace{0.2cm} Library: {\scriptsize\url{https://github.com/erodola/nodmod}}}; we retained genre-diverse songs with large acoustic variability, whose 3.7k samples are shorter than 2s. Table~\ref{tab:dataset-statistics} summarizes the corpus.


\subsection{Music transfer}
Listening examples, with the corresponding LAC text descriptions, are available online.\footnote{Demo page: {\scriptsize\url{https://erodola.github.io/lac-demo/}}}
We encourage readers to listen to them, as they give a clear sense of the quality of the reconstructions.

For song-level transfer, we separate {symbolic structure} from {sound}. The symbolic channel preserves the musical content (notes, timing, and patterns) while LAC carries the acoustic character of the instruments. Tracker modules make this separation natural: their patterns remain symbolic, and their embedded instrument samples can be replaced independently.
%
%
More generally, MIDI or another score-like format could serve the same symbolic role.

\begin{figure}[t]
    \centering
    \begin{overpic}[height=0.225\linewidth,trim=0 -0.25cm 0 0]{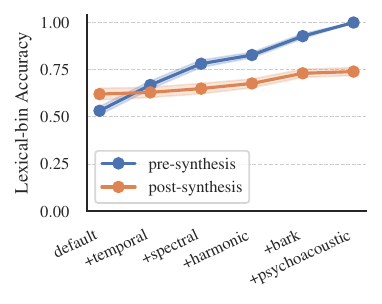}
        \put(55,0){\scriptsize(a)}
    \end{overpic}
    \begin{overpic}[height=0.225\linewidth,trim=0 -0.25cm 0 0]{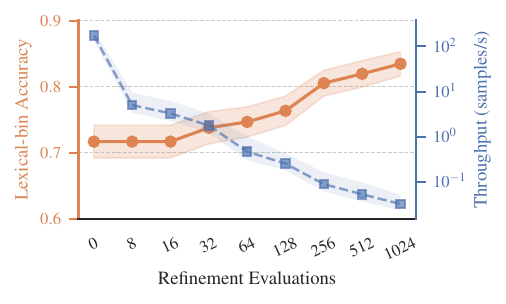}
        \put(46,0){\scriptsize(b)}
    \end{overpic}
    \begin{overpic}[width=0.33\linewidth,trim=0 -0.8cm 0 0]{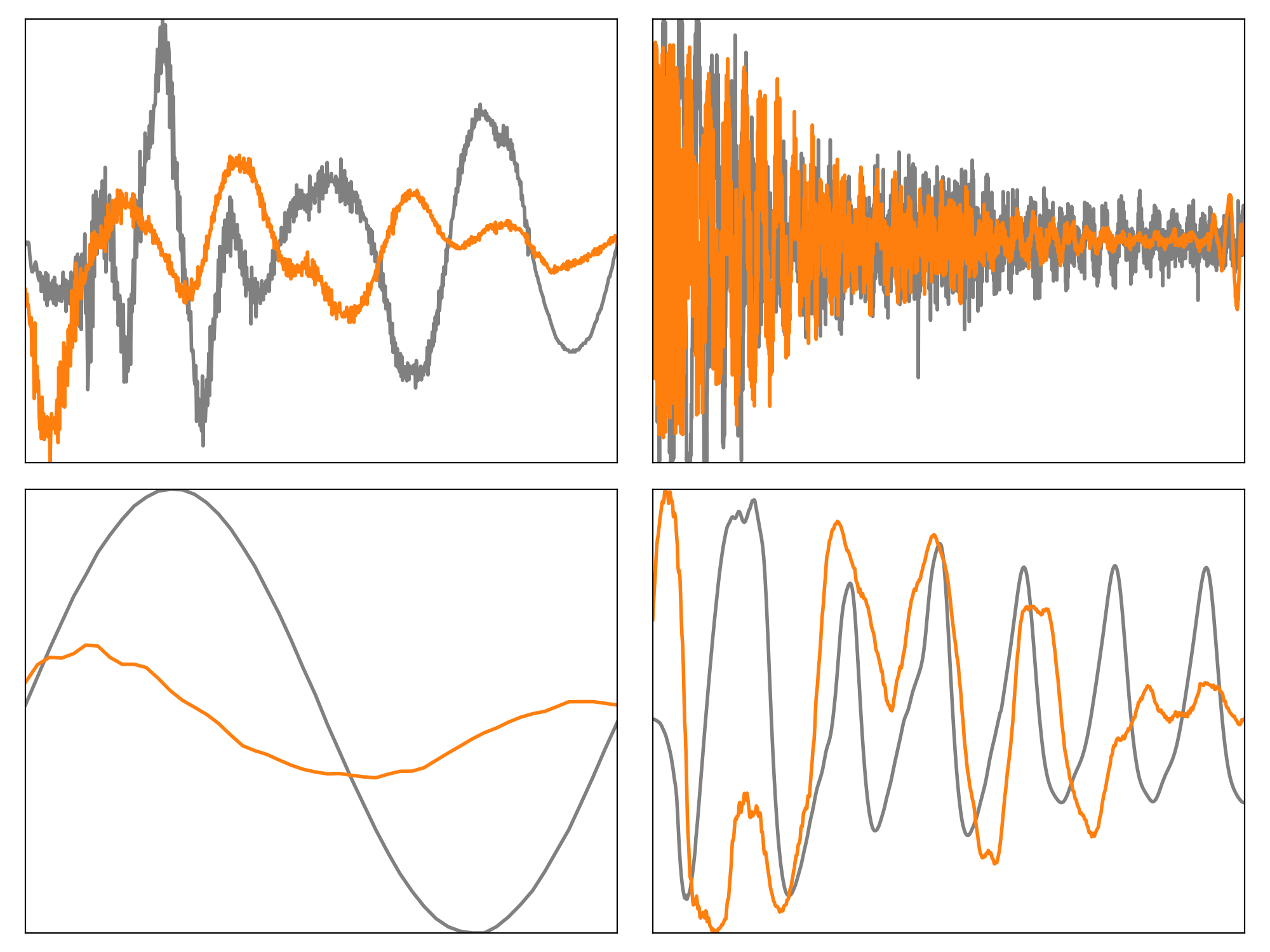}
        \put(48,0){\scriptsize(c)}
        \put(27.5,72){\scriptsize kickdrum}
        \put(85,72){\scriptsize snare}
        \put(20,35){\scriptsize single period}
        \put(71,35){\scriptsize electric bass}
    \end{overpic}
    \caption{\textbf{Feature-family and refinement analyses, plus qualitative waveform examples.} (a) Lexical-bin accuracy as feature families are added cumulatively, measured both before rendering and after full synthesis. (b) Post-synthesis lexical-bin accuracy and throughput as the number of closed-loop refinement evaluations increases. (c) Four representative original waveforms ({\color{gray}\textbf{gray}}) and LAC reconstructions ({\color{orange}\textbf{orange}}). The instrument labels are approximate. Exact sample-level agreement is loose, but the main envelope and periodic structure are preserved, contributing to perceptual similarity. Corresponding audio can be heard on the demo page (samples 03, 73, 50, and 45).}
    \label{fig:feature-group-reconstruction}
\end{figure}

\subsection{Feature-family ablation} \label{sec:feature-family-ablation}
To assess which parts of the lexical code matter most, we add feature families cumulatively and measure lexical-bin reconstruction accuracy after each step (Figure~\ref{fig:feature-group-reconstruction}a). We report two views. In \emph{pre-synthesis} evaluation, labels are inverted to representative feature values before any waveform is rendered. In \emph{post-synthesis}, the decoded targets are rendered to audio and features are re-extracted from the result. Shaded bands are 95\% CIs (${\pm}1.96\,\mathrm{SEM}$, Normal approximation, $n{=}100$).

Pre-synthesis accuracy rises steadily and reaches 100\% once all feature families are included, confirming that the encode--transmit--decode chain preserves the full lexical description. Post-synthesis is harder: the renderer must produce a waveform whose re-extracted features fall back into the intended bins. Here, Bark-band information provides the largest gain, while temporal and psychoacoustic families add little or no improvement. With all groups included, post-synthesis accuracy reaches about 74\%, compared with the 100\% pre-synthesis ceiling. The gap reflects a renderer limitation rather than an information loss in the textual channel.

\subsection{Closed-loop refinement}
Closed-loop refinement improves reconstruction by repeatedly re-rendering candidate waveforms and scoring their re-extracted features. Figure~\ref{fig:feature-group-reconstruction}b, plots post-synthesis lexical-bin accuracy against the evaluation budget (error bands as above). Little changes up to 16 evaluations. Gains appear at 32 evaluations, are strongest between 128 and 256, and then continue steadily. Overall, refinement raises accuracy from roughly 72\% without refinement to roughly 84\%, but at a corresponding runtime cost that decreases throughput approximately inversely with the number of evaluations. 
We used 64 evaluations in our other experiments as a reasonable tradeoff between accuracy and runtime.

Figure~\ref{fig:feature-group-reconstruction}c, provides a qualitative view of the same behavior. The reconstructions are not expected to match the originals pointwise: LAC is a generative acoustic channel, not a waveform codec. Even when the traces differ visibly, the reconstructions often preserve the macroscopic envelope, dominant periodicity, and tonal-versus-noisy balance that drive perceived similarity. This is best judged by listening; the corresponding examples are available on the demo page.

\section{Limitations}
\label{sec:limitations}
This work has several limitations.
First, the system targets short, isolated, non-speech sounds such as hits, bursts, plucks, and short notes. It is not a speech, music, or general-purpose audio codec. Its current 47-coordinate representation captures global acoustic character, but not phonetic content, speaker identity, lyrics, melody, harmony, meter, arrangement, or other long-range structure. Sustained and time-varying sounds therefore remain difficult; dedicated tests on overlapping-window descriptions proved more fragile without preserving coherent temporal continuity.

Second, decoding is deliberately approximate by design. Labels invert to representative intervals rather than exact measurements. LAC therefore reconstructs sounds consistent with a lexical acoustic description, rather than recovering the original waveform exactly.

Finally, symbolic music transfer is only partially developed. We currently pass symbolic structure in ABC notation, which is practical but not a general solution: describing pitch, rhythm, meter, voicing, repetition, and form in unconstrained prose quickly becomes a constraint-satisfaction (SAT-like) problem. We therefore treat this as a pragmatic extension, not as evidence that arbitrary symbolic music can already be transmitted through natural language.

\paragraph{Broader impact and risks.}
A language-native acoustic representation could make sound synthesis and transfer more interpretable, inspectable, and controllable. Because the lexical representation is human-readable, users can audit which aspects of a sound are being preserved or altered, which is harder to do with opaque learned latents. Transferring symbolic note patterns also suggests a practical route for controllable music and sound-design workflows in which structure and timbre are manipulated separately.

There are also risks. Any system that makes audio easier to describe, transfer, and reconstruct can be repurposed for imitation or spoofing, particularly if the language interface becomes more powerful. 

\section{Conclusion}
We introduced lexical acoustic coding (LAC), a framework for communicating sound through natural language. LAC converts audio into interpretable acoustic descriptors, quantizes them into a compact lexical code, verbalizes the code as structured English, and reconstructs audio from interval-valued acoustic constraints. This places LAC between captions and codecs: more structured than free-form text, but more readable than learned latents.
Experiments show that lexical descriptions preserve meaningful acoustic information through an agent-mediated text channel. We also demonstrate symbolic music transfer, where LAC carries timbre while ABC notation carries musical structure.

LAC is not currently designed for exact waveform recovery, speech coding, or full music compression. It reconstructs sounds consistent with a transmitted description. Future work should address time-varying audio, improve decoding, refine the vocabulary, and evaluate with broader listening tests.

\bibliographystyle{plainnat}
\bibliography{references}

\appendix

\section{Comparison table: axis definitions and per-method justifications}
\label{app:comparison-table}
We expand on Table~\ref{tab:lac_comparison}. We first give a precise definition of each axis, then justify, method by method, the entry assigned in every cell. 

\subsection{Axes}
\begin{description}
\item[Human readability.] Whether the representation, as transmitted, is directly intelligible to a human reader without any decoding software.

\item[LLM-native transport.] Whether the representation can be natively consumed and produced by general-purpose text LLMs.

\item[Semantic editing.] Whether a user or agent can edit the representation through high-level semantic operations, such as replacing a word, adjusting an attribute, and have the change propagate to the reconstructed audio in a predictable way. 

\item[Acoustic interpretability.] Whether the components of the representation correspond to known acoustic or perceptual quantities, allowing a domain expert to read them off acoustic properties from the representation directly, without training a probe.

\item[Training-free.] Whether the representation itself can be obtained from the audio without the use of any machine learning model. This axis concerns only the production of the representation, not how it is later transported, modeled, or rendered back to audio.

\item[Generative decoding.] Whether the decoder acts as a model over the representation space, capable of synthesizing plausible audio for arbitrary representation values rather than merely inverting recordings it was given. 

\item[Bandwidth efficiency.] Whether the representation is compact relative to the perceptual content it carries.

\item[Bit-wise reconstruction.] Whether the representation allows recovery of the original waveform exactly, sample-for-sample.
\end{description}

\subsection{Per-method justification}
\paragraph{Lossless codec (FLAC, WAV).}
The transmitted form is a binary stream, opaque to humans (\emph{Human readability}: no) and to text LLMs (\emph{LLM-native transport}: no). The representation supports no semantic operations on its content (\emph{Semantic editing}: no) and exposes no acoustic descriptors (\emph{Acoustic interpretability}: no). Encoding and decoding are deterministic algorithms with no learned components (\emph{Training-free}: yes). \emph{Generative decoding}: no, since the decoder only inverts the encoded form: arbitrary bitstream values do not correspond to coherent audio, so the decoder is not a model over the representation space. It compresses far above the rates achievable by neural codecs at comparable quality, so we score \emph{Bandwidth efficiency}: no relative to the rest of the table. The defining property of the row is exact recovery (\emph{Bit-wise reconstruction}: yes).

\paragraph{Handcrafted descriptors (MFCC, spectral centroid).}
This row stands for handcrafted feature-vector representations as a class, including spectral centroid, brightness, roll-off, harmonicity, MFCC, and similar descriptors. The transmitted form is a floating-point vector, not directly readable (\emph{Human readability}: no), not text-tokenizable (\emph{LLM-native transport}: no), and not editable through semantic operations (\emph{Semantic editing}: no). We score \emph{Acoustic interpretability}: yes because the components correspond to known acoustic and perceptual quantities; we note that higher MFCC coefficients are themselves DCT components and only weakly interpretable in isolation, but the row as a whole is dominated by physically-meaningful descriptors. Extraction is deterministic and uses no learned components (\emph{Training-free}: yes). \emph{Generative decoding}: no, since there is no standard decoder that acts as a model over the descriptor space; feature-to-spectrogram approximations recover one specific spectrum rather than synthesizing plausible audio for arbitrary descriptor values. The vectors are small per second of audio (\emph{Bandwidth efficiency}: yes) but discard phase and most signal detail (\emph{Bit-wise reconstruction}: no).

\paragraph{Neural codec (EnCodec, SoundStream).}
Discrete latent tokens are not human-readable (\emph{Human readability}: no). We assign \emph{LLM-native transport}: partial because these tokens are widely consumed by audio language models such as AudioLM \citep{borsos2022_audiolm} and MusicLM \citep{agostinelli2023_musiclm}, so they do flow through language-style sequence models; however, those models rely on codec-specific token vocabularies and audio-side modeling rather than general-purpose text LLMs operating on the stream as ordinary text. The tokens expose no semantic edit interface (\emph{Semantic editing}: no) and no interpretable acoustic axes (\emph{Acoustic interpretability}: no). Encoder and decoder are jointly trained (\emph{Training-free}: no). \emph{Generative decoding}: yes, because the neural decoder is trained as a model over the latent space and synthesizes plausible audio for arbitrary token sequences, including modified or sampled ones. The codec operates at low bitrate (\emph{Bandwidth efficiency}: yes), without exact recovery (\emph{Bit-wise reconstruction}: no).

\paragraph{Audio-language tokenizers (AudioLM, TASTE, TaDiCodec, TADA, UniAudio~2.0).}
The tokens themselves are not human-readable (\emph{Human readability}: no). We assign \emph{LLM-native transport}: partial because these tokenizations are designed for joint modeling with text by language-model-style architectures (often via text-alignment, interleaving, or hierarchical semantic/acoustic decompositions), which moves them closer to text-LM territory than codec tokens. They still fall short of native text transport, however: the tokens live in a custom audio vocabulary and require a model that has been trained or fine-tuned on that vocabulary; a pretrained general-purpose text LLM cannot consume them as-is. \emph{Semantic editing}: partial applies only to the text-aligned subset of this family (TASTE \citep{tseng2025_taste}, TADA \citep{dang2026_tada}, TaDiCodec \citep{wang2025_tadicodec}). The audio tokens themselves remain opaque integers and cannot be edited directly; editing operates on the text that conditions or accompanies the tokens, and the decoder rerenders audio under the modified text. Methods without text conditioning (AudioLM \citep{borsos2022_audiolm}, vanilla acoustic-token branches such as those in UniAudio~2.0 \citep{yang2026_uniaudio2}) do not support semantic editing at all. The latents are learned and not perceptually interpretable (\emph{Acoustic interpretability}: no), require training (\emph{Training-free}: no), and pair with neural decoders that act as models over the token space, synthesizing plausible audio for arbitrary token sequences (\emph{Generative decoding}: yes), at low rates (\emph{Bandwidth efficiency}: yes), without exact recovery (\emph{Bit-wise reconstruction}: no).

\paragraph{Unconstrained text caption.}
A caption is text (\emph{Human readability}: yes), natively handled by LLMs (\emph{LLM-native transport}: yes), and semantically editable by construction (\emph{Semantic editing}: yes). \emph{Acoustic interpretability}: partial acknowledges that captions describe sounds at a semantic rather than acoustic level (``a dog barks in a hallway'' identifies source and scene but does not commit to spectral content), so the representation is interpretable in a weaker sense than a descriptor vector. We score \emph{Training-free}: yes because a caption is a piece of text that can be produced without any model (by a human listener, for example), so the representation itself does not require a learned encoder, even though captioners are commonly used in practice. \emph{Generative decoding}: yes, since the text-to-audio decoder is a generative model over the caption space and synthesizes plausible audio for arbitrary captions. Captions can be longer than the perceptual content they convey, especially when a single descriptor would suffice (\emph{Bandwidth efficiency}: no), and recovery is not exact (\emph{Bit-wise reconstruction}: no).

\paragraph{LAC (this paper).}
LAC transmits a constrained natural-language sentence over classical acoustic descriptors. The sentence is text, hence \emph{Human readability}: yes and \emph{LLM-native transport}: yes. \emph{Semantic editing}: yes because the sentence is parsed into the descriptor vector through a fixed vocabulary mapping, so an edit to a word induces a predictable edit on the descriptor and on the synthesized audio. The descriptors are physically meaningful by construction (\emph{Acoustic interpretability}: yes). \emph{Training-free}: yes because the representation can be obtained by deterministic feature extraction followed by rule-based lexicalization and a human turning it into prose, with no learned encoder; in practice we use an LLM to phrase the sentence, but this is a convenience rather than a requirement, mirroring the caption case. \emph{Generative decoding}: yes because the synthesizer renders plausible audio for arbitrary descriptor vectors, acting as a model over the representation space; the synthesizer itself is deterministic given the descriptors, but its role is generative as the representation supports synthesis of arbitrary content, not just inversion of recorded points. Sentences are longer per second of audio than neural-codec tokens but substantially shorter than free-form captions, so we score \emph{Bandwidth efficiency}: partial. Reconstruction is necessarily lossy (\emph{Bit-wise reconstruction}: no).

\section{Feature inventory}
\label{app:feature-list}
Table~\ref{tab:feature64} reports all 47 features in canonical extraction order, with concise operational definitions.

\begingroup
\renewcommand{\feat}[1]{\textcolor{lacFeat}{\texttt{#1}}}
\setlength{\LTleft}{0pt}
\setlength{\LTright}{0pt}
\setlength{\tabcolsep}{5pt}

\small
\begin{longtable}{>{\centering\arraybackslash}p{0.05\textwidth}>{\raggedright\arraybackslash}p{0.25\textwidth}>{\raggedright\arraybackslash}p{0.60\textwidth}}
\caption{\textbf{Type legend.} \texttt{T}: Temporal; \texttt{S}: Spectral; \texttt{H}: Harmonic; \texttt{B}: Psychoacoustic Bark-band; \texttt{N}: Psychoacoustic non-Bark.}
\label{tab:feature64}\\
\toprule
\textbf{Type} & \textbf{Feature} & \textbf{What it captures} \\
\midrule
\endfirsthead
\toprule
\textbf{Type} & \textbf{Feature} & \textbf{What it captures} \\
\midrule
\endhead
\midrule
\multicolumn{3}{r}{\footnotesize Continued on next page}\\
\endfoot
\bottomrule
\endlastfoot
\texttt{T} & \feat{rms\_energy} & Root-mean-square of waveform samples; overall energy. \\
\texttt{T} & \feat{crest\_factor\_db} & $20 \log_{10}(\text{peak}/\text{RMS})$; transient peakiness vs average level. \\
\texttt{T} & \feat{zero\_crossing\_rate} & Zero crossings per second (sign changes / duration). \\
\texttt{T} & \feat{log\_attack\_time} & $\log_{10}$ time for smoothed envelope to rise from 20\% to 90\% of peak. \\
\texttt{T} & \feat{attack\_slope\_db\_s} & Attack slope in dB/s between 10\% and 90\% of peak envelope. \\
\texttt{T} & \feat{temporal\_centroid} & Energy centroid of frame RMS along time, normalized by duration. \\
\texttt{T} & \feat{decay\_time\_s} & Exponential decay time constant from log-envelope regression after peak. \\
\texttt{S} & \feat{spectral\_centroid\_hz} & Magnitude-weighted mean frequency (Hz), averaged over frames. \\
\texttt{S} & \feat{spectral\_flatness} & Geometric/arithmetic mean of power spectrum (Wiener entropy), avg. over frames. \\
\texttt{S} & \feat{spectral\_rolloff\_hz} & Frequency below which 85\% of spectral energy lies, averaged over frames. \\
\texttt{S} & \feat{spectral\_flux} & Mean squared positive change between successive magnitude spectra. \\
\texttt{S} & \feat{spectral\_kurtosis} & Fourth standardized moment of mean magnitude spectrum. \\
\texttt{S} & \feat{spectral\_entropy} & Normalized Shannon entropy of power spectrum (mean spectrum). \\
\texttt{S} & \feat{spectral\_irregularity} & Jensen irregularity: sum of squared adjacent-bin diffs / total squared mag. \\
\texttt{H} & \feat{f0\_hz} & Fundamental frequency via YIN; median of per-frame estimates. \\
\texttt{H} & \feat{harmonic\_noise\_ratio\_db} & HNR in dB from normalized autocorrelation peak, avg. over frames. \\
\texttt{H} & \feat{inharmonicity} & Amplitude-weighted avg. relative deviation of partials from $k \cdot f_0$. \\
\texttt{H} & \feat{tristimulus\_1} & Energy ratio of harmonic 1 to total harmonic energy. \\
\texttt{H} & \feat{tristimulus\_2} & Energy ratio of harmonics 2-4 to total harmonic energy. \\
\texttt{H} & \feat{tristimulus\_3} & Energy ratio of harmonics 5+ to total harmonic energy. \\
\texttt{H} & \feat{odd\_even\_harmonic\_ratio} & Ratio of odd-harmonic energy to even-harmonic energy. \\
\texttt{B} & \feat{bark\_band\_1} & Log(1+band power) in 20-100 Hz critical band. \\
\texttt{B} & \feat{bark\_band\_2} & Log(1+band power) in 100-200 Hz critical band. \\
\texttt{B} & \feat{bark\_band\_3} & Log(1+band power) in 200-300 Hz critical band. \\
\texttt{B} & \feat{bark\_band\_4} & Log(1+band power) in 300-400 Hz critical band. \\
\texttt{B} & \feat{bark\_band\_5} & Log(1+band power) in 400-510 Hz critical band. \\
\texttt{B} & \feat{bark\_band\_6} & Log(1+band power) in 510-630 Hz critical band. \\
\texttt{B} & \feat{bark\_band\_7} & Log(1+band power) in 630-770 Hz critical band. \\
\texttt{B} & \feat{bark\_band\_8} & Log(1+band power) in 770-920 Hz critical band. \\
\texttt{B} & \feat{bark\_band\_9} & Log(1+band power) in 920-1080 Hz critical band. \\
\texttt{B} & \feat{bark\_band\_10} & Log(1+band power) in 1080-1270 Hz critical band. \\
\texttt{B} & \feat{bark\_band\_11} & Log(1+band power) in 1270-1480 Hz critical band. \\
\texttt{B} & \feat{bark\_band\_12} & Log(1+band power) in 1480-1720 Hz critical band. \\
\texttt{B} & \feat{bark\_band\_13} & Log(1+band power) in 1720-2000 Hz critical band. \\
\texttt{B} & \feat{bark\_band\_14} & Log(1+band power) in 2000-2320 Hz critical band. \\
\texttt{B} & \feat{bark\_band\_15} & Log(1+band power) in 2320-2700 Hz critical band. \\
\texttt{B} & \feat{bark\_band\_16} & Log(1+band power) in 2700-3150 Hz critical band. \\
\texttt{B} & \feat{bark\_band\_17} & Log(1+band power) in 3150-3700 Hz critical band. \\
\texttt{B} & \feat{bark\_band\_18} & Log(1+band power) in 3700-4400 Hz critical band. \\
\texttt{B} & \feat{bark\_band\_19} & Log(1+band power) in 4400-5300 Hz critical band. \\
\texttt{B} & \feat{bark\_band\_20} & Log(1+band power) in 5300-6400 Hz critical band. \\
\texttt{B} & \feat{bark\_band\_21} & Log(1+band power) in 6400-7700 Hz critical band. \\
\texttt{B} & \feat{bark\_band\_22} & Log(1+band power) in 7700-9500 Hz critical band. \\
\texttt{B} & \feat{bark\_band\_23} & Log(1+band power) in 9500-12000 Hz critical band. \\
\texttt{B} & \feat{bark\_band\_24} & Log(1+band power) in 12000-15500 Hz critical band. \\
\texttt{N} & \feat{sharpness\_acum} & Zwicker sharpness (acum) using DIN 45692 g(z) and Bark-band $E^{0.23}$. \\
\texttt{N} & \feat{roughness} & Vassilakis roughness from pairwise peak interactions (Plomp-Levelt curve). \\
\end{longtable}
\endgroup
\section{Feature provenance}\label{app:feature-provenance}
This appendix documents the provenance of the 47 acoustic descriptors used in the LAC feature extractor. Table \ref{tab:feature-provenance} groups each feature by its closest reference implementation or off-the-shelf analogue, showing that the lexical code is grounded in established DSP descriptors rather than learned latent variables.

\begin{longtable}{p{0.30\linewidth}p{0.62\linewidth}}
\caption{Reference implementation of the 47 acoustic descriptors.}
\label{tab:feature-provenance}\\
\toprule
\textbf{Feature} & \textbf{Reference / off-the-shelf analogue} \\
\midrule
\endfirsthead

\toprule
\textbf{Feature} & \textbf{Reference / off-the-shelf analogue} \\
\midrule
\endhead

\midrule
\multicolumn{2}{r}{\emph{Continued on next page}}\\
\endfoot

\bottomrule
\endlastfoot

\feat{rms\_energy}, \feat{zero\_crossing\_rate}, \feat{spectral\_centroid\_hz}, \feat{spectral\_rolloff\_hz} &
librosa \citep{mcfee2015_librosa} implementation of standard DSP features \\
\addlinespace[2pt]

\feat{crest\_factor\_db}, \feat{log\_attack\_time}, \feat{attack\_slope\_db\_s}, \feat{temporal\_centroid}, \feat{spectral\_kurtosis}, \feat{spectral\_flux}, \feat{odd\_even\_harmonic\_ratio}, \feat{inharmonicity} &
Timbre Toolbox \citep{peeters2011timbretoolbox} implementation of MPEG--7 \citep{iso15938_4_2002} and CUIDADO \citep{peeters2004large} features \\
\addlinespace[2pt]

\feat{decay\_time\_s} &
Adapted from standard decay/envelope fitting; related to Timbre Toolbox temporal decrease descriptors \citep{peeters2011timbretoolbox} \\
\addlinespace[2pt]

\feat{spectral\_flatness} &
\citep{dubnov2004flatness,peeters2011timbretoolbox}; also in librosa \\
\addlinespace[2pt]

\feat{spectral\_entropy} &
Eq. (2) in \citep{misra2004spectralentropy} \\
\addlinespace[2pt]

\feat{spectral\_irregularity} &
Spectral irregularity / timbre-model descriptor \citep{jensen1999timbre,peeters2011timbretoolbox} \\
\addlinespace[2pt]

\feat{f0\_hz} &
\citep{decheveigne2002yin}; also in librosa \\
\addlinespace[2pt]

\feat{harmonic\_noise\_ratio\_db} &
Eq. (4) in \citep{boersma1993hnr} \\
\addlinespace[2pt]

\feat{tristimulus\_1..3} &
\citep{pollardJansson1982,peeters2011timbretoolbox} \\
\addlinespace[2pt]

\feat{bark\_band\_1..24} &
Bark critical bands \citep{fastlZwicker2007} \\
\addlinespace[2pt]

\feat{sharpness\_acum} &
DIN/Zwicker-style sharpness \citep{din45692_2009,fastlZwicker2007,halesSwiftGee2017sharpness} \\
\addlinespace[2pt]

\feat{roughness} &
\citep{plompLevelt1965,vassilakis2001roughness,virtanen2020scipy} \\

\end{longtable}

\section{Full Feature Vocabulary Mapping}
\label{app:feature-vocab-mapping}

This appendix includes the full feature-to-vocabulary mapping used by the lexical encoder, as well as a detailed specification of the backward $R_i:\mathcal{A}_i\to\mathbb{R}$ maps.

\subsection{Representative values for lexical inversion}
\label{app:representative-values}

At decode time, each lexical label is used in two different ways. First, it defines the interval associated with the transmitted lexical bin; that interval is retained for constraint checking during refinement. Second, it is mapped to a single representative numeric value used to initialize or directly set synthesis parameters. These two roles should not be conflated: the interval is the target set, whereas the representative is only a deterministic anchor inside or near that set.

For a label whose interval is $[l,u)$, the generic representative-value rule applies:
\begin{equation}
\operatorname{rep}(l,u)=
\begin{cases}
\mathrm{NaN}, & l=u=\mathrm{NaN},\\[4pt]
\dfrac{l+u}{2}, & l,u\in\mathbb{R}, \quad \text{(midpoint)}\\[8pt]
1.5\,l, & u=+\infty,\; l>0,\\[4pt]
1.0, & u=+\infty,\; l=0,\\[4pt]
1.5\,|l|+1.0, & u=+\infty,\; l<0,\\[4pt]
0.5\,u, & l=-\infty,\; u>0,\\[4pt]
u-\dfrac{|u|}{2}-1.0, & l=-\infty,\; u\le 0.
\end{cases}
\label{eq:generic-representative}
\end{equation}

Equation~\eqref{eq:generic-representative} is used for $245$ out of $285$ bins, excluding the \feat{f0\_hz} feature which uses a geometric mean, being log-spaced (see below).

For finite intervals, Eq.~\eqref{eq:generic-representative} computes the ordinary arithmetic midpoint. 

For open-ended intervals, however, there is no true midpoint, so the decoder uses a simple deterministic heuristic instead. These open-ended representatives are synthesis anchors that work well in practice.

Sentinel labels that represent undefined or structurally absent values are mapped to $\mathrm{NaN}$ rather than to any finite number. In the current vocabulary this includes, for example, labels such as \labeltag{unpitched}, \labeltag{onset-undetected}, \labeltag{slope-undefined}, and \labeltag{non-decaying}. This is important because the decoder distinguishes between a finite target and the explicit absence of that target.

\paragraph{Bark-band labels.}
Bark-band labels are composite strings such as \labeltag{dominant rumble} or \labeltag{present air}. For these features, the band identity is carried by the feature name itself (\feat{bark\_band\_1}, \dots, \feat{bark\_band\_24}), while the representative value is determined only by the first word of the lexical label, i.e., the level prefix (\labeltag{silent}, \labeltag{trace}, \labeltag{faint}, \labeltag{present}, \labeltag{strong}, \labeltag{dominant}, \labeltag{overwhelming}). As a result, \labeltag{dominant rumble} and \labeltag{dominant air} share the same numeric representative; what differs between them is the Bark-band index, not the level value itself.

Applying Eq.~\eqref{eq:generic-representative} to the finite Bark levels yields the representatives
\begin{center}
\labeltag{silent}$\mapsto$0.005,
\labeltag{trace}$\mapsto$1.005,
\labeltag{faint}$\mapsto$3.5,
\labeltag{present}$\mapsto$6.5,
\labeltag{strong}$\mapsto$9.5,
\labeltag{dominant}$\mapsto$13.0.
\end{center}
The open-ended Bark level \labeltag{overwhelming} would generically map to $1.5\times 15 = 22.5$, but for better qualitative results we override this and use
\begin{center}
\labeltag{overwhelming} $\mapsto$ 18.0
\end{center}
instead, to keep the Bark targets in a more conservative range for synthesis.

\paragraph{Special treatment of \feat{f0\_hz}.}
Fundamental-frequency labels are treated differently from ordinary finite bins because the pitch vocabulary is logarithmically spaced. For any finite positive \feat{f0\_hz} interval $[l,u)$, the representative value is taken to be the geometric mean
\begin{equation}
\operatorname{rep}_{f_0}(l,u)=\sqrt{lu},
\label{eq:f0-geometric-mean}
\end{equation}
rather than the arithmetic midpoint. This places the representative at the center of the bin on the log-frequency axis, which is the natural geometry of the pitch partition. If an \feat{f0\_hz} interval touches zero or is open-ended, the implementation falls back to the generic rule in Eq.~\eqref{eq:generic-representative}. The sentinel \labeltag{unpitched} label maps to $\mathrm{NaN}$.

\paragraph{Overrides.}
After the generic representative-value rule is applied, a small number of labels are replaced with hard-coded representatives. These overrides act as post-processing adjustments used to keep open-ended or extreme labels within a numerically and physically reasonable range for synthesis. Table~\ref{tab:representative-overrides} lists all explicit overrides used by the decoder.

\begin{table}[t]

  \caption{Representative-value overrides. These hand-selected values are used only for labels whose generic inversion value would be uninformative or implausible.}
  \label{tab:representative-overrides}

  \centering
  \small
  \setlength{\tabcolsep}{5pt}
  \begin{tabular}{@{}llr@{}}
    \toprule
    Feature & Label & Value \\
    \midrule
    \feat{rms\_energy} & \labeltag{thunderous} & 0.75 \\
    \feat{crest\_factor\_db} & \labeltag{spiky} & 22.0 \\
    \feat{zero\_crossing\_rate} & \labeltag{extreme-oscillation} & 15{,}000.0 \\
    \feat{log\_attack\_time} & \labeltag{snap-onset} & $-3.2$ \\
    \feat{log\_attack\_time} & \labeltag{creeping-onset} & $-1.0$ \\
    \feat{attack\_slope\_db\_s} & \labeltag{explosive} & 18{,}000.0 \\
    \feat{decay\_time\_s} & \labeltag{endless} & 12.0 \\
    \feat{spectral\_centroid\_hz} & \labeltag{sizzling} & 14{,}000.0 \\
    \feat{spectral\_rolloff\_hz} & \labeltag{open-ceiling} & 16{,}000.0 \\
    \feat{spectral\_spread\_hz} & \labeltag{ultra-wide} & 7{,}000.0 \\
    \feat{spectral\_skewness} & \labeltag{extreme-asymmetry} & 75.0 \\
    \feat{spectral\_kurtosis} & \labeltag{needle-point} & 5{,}000.0 \\
    \feat{spectral\_slope} & \labeltag{ascending} & 0.00005 \\
    \feat{harmonic\_noise\_ratio\_db} & \labeltag{pristine} & 20.0 \\
    \feat{harmonic\_noise\_ratio\_db} & \labeltag{noise-engulfed} & $-5.0$ \\
    \feat{inharmonicity} & \labeltag{warped} & 0.15 \\
    \feat{odd\_even\_harmonic\_ratio} & \labeltag{fundamentals-only} & 75.0 \\
    \feat{sharpness\_acum} & \labeltag{piercing} & 6.0 \\
    \feat{roughness} & \labeltag{abrasive} & 0.85 \\
    \bottomrule
  \end{tabular}

\end{table}

In summary, the decoder does not uniformly use literal midpoints. For finite bins it uses arithmetic midpoints; for open-ended bins it uses deterministic heuristic representatives; for finite positive \feat{f0\_hz} bins it uses geometric means; and for a small number of extreme labels it applies hand-chosen overrides. The interval itself remains the authoritative lexical constraint during refinement, while the representative value serves only as a stable numeric anchor for synthesis.

\subsection{Full lexical code}

\feat{rms\_energy}

\vspace{-1ex}
\hspace{0.3cm} $[0.0, 0.02)$: \labeltag{whisper}

\vspace{-1ex}
\hspace{0.3cm} $[0.02, 0.1)$: \labeltag{hushed}

\vspace{-1ex}
\hspace{0.3cm} $[0.1, 0.3)$: \labeltag{mid-power}
  
\vspace{-1ex}
\hspace{0.3cm} $[0.3, 0.55)$: \labeltag{forceful}
  
\vspace{-1ex}
\hspace{0.3cm} $[0.55, \infty)$: \labeltag{thunderous}
  
\feat{crest\_factor\_db}

\vspace{-1ex}
\hspace{0.3cm} $[0.0, 5.0)$: \labeltag{sustained}

\vspace{-1ex}
\hspace{0.3cm} $[5.0, 10.0)$: \labeltag{rounded}

\vspace{-1ex}
\hspace{0.3cm} $[10.0, 14.0)$: \labeltag{punchy}
  
\vspace{-1ex}
\hspace{0.3cm} $[14.0, 17.0)$: \labeltag{impulsive}
  
\vspace{-1ex}
\hspace{0.3cm} $[17.0, \infty)$: \labeltag{spiky}

\feat{zero\_crossing\_rate}

\vspace{-1ex}
\hspace{0.3cm} $[0.0, 100.0)$: \labeltag{infrasonic}

\vspace{-1ex}
\hspace{0.3cm} $[100.0, 500.0)$: \labeltag{low-oscillation}

\vspace{-1ex}
\hspace{0.3cm} $[500.0, 2000.0)$: \labeltag{mid-oscillation}
  
\vspace{-1ex}
\hspace{0.3cm} $[2000.0, 10000.0)$: \labeltag{high-oscillation}
  
\vspace{-1ex}
\hspace{0.3cm} $[10000.0, \infty)$: \labeltag{extreme-oscillation}

\feat{log\_attack\_time}

\vspace{-1ex}
\hspace{0.3cm} $\mathrm{NaN}$: \labeltag{onset-undetected}

\vspace{-1ex}
\hspace{0.3cm} $[-\infty, -2.8)$: \labeltag{snap-onset}

\vspace{-1ex}
\hspace{0.3cm} $[-2.8, -2.5)$: \labeltag{swift-onset}

\vspace{-1ex}
\hspace{0.3cm} $[-2.5, -2.0)$: \labeltag{moderate-onset}
  
\vspace{-1ex}
\hspace{0.3cm} $[-2.0, -1.5)$: \labeltag{gradual-onset}
  
\vspace{-1ex}
\hspace{0.3cm} $[-1.5, \infty)$: \labeltag{creeping-onset}

\feat{attack\_slope\_db\_s}

\vspace{-1ex}
\hspace{0.3cm} $\mathrm{NaN}$: \labeltag{slope-undefined}

\vspace{-1ex}
\hspace{0.3cm} $[0.0, 3000.0)$: \labeltag{feathered}

\vspace{-1ex}
\hspace{0.3cm} $[3000.0, 8000.0)$: \labeltag{measured}

\vspace{-1ex}
\hspace{0.3cm} $[8000.0, 14000.0)$: \labeltag{aggressive}
  
\vspace{-1ex}
\hspace{0.3cm} $[14000.0, \infty)$: \labeltag{explosive}

\feat{temporal\_centroid}

\vspace{-1ex}
\hspace{0.3cm} $[0.0, 0.15)$: \labeltag{front-loaded}

\vspace{-1ex}
\hspace{0.3cm} $[0.15, 0.25)$: \labeltag{front-weighted}

\vspace{-1ex}
\hspace{0.3cm} $[0.25, 0.4)$: \labeltag{centered}
  
\vspace{-1ex}
\hspace{0.3cm} $[0.4, 0.55)$: \labeltag{evenly-distributed}
  
\vspace{-1ex}
\hspace{0.3cm} $[0.55, 1.0)$: \labeltag{back-loaded}

\feat{decay\_time\_s}

\vspace{-1ex}
\hspace{0.3cm} $\mathrm{NaN}$: \labeltag{non-decaying}

\vspace{-1ex}
\hspace{0.3cm} $[0.0, 0.04)$: \labeltag{clipped}

\vspace{-1ex}
\hspace{0.3cm} $[0.04, 0.12)$: \labeltag{staccato}

\vspace{-1ex}
\hspace{0.3cm} $[0.12, 0.4)$: \labeltag{short-decay}
  
\vspace{-1ex}
\hspace{0.3cm} $[0.4, 2.0)$: \labeltag{lingering}
  
\vspace{-1ex}
\hspace{0.3cm} $[2.0, 10.0)$: \labeltag{ringing}

\vspace{-1ex}
\hspace{0.3cm} $[10.0, \infty)$: \labeltag{endless}

\feat{spectral\_centroid\_hz}

\vspace{-1ex}
\hspace{0.3cm} $[0.0, 150.0)$: \labeltag{subterranean}

\vspace{-1ex}
\hspace{0.3cm} $[150.0, 500.0)$: \labeltag{dark}

\vspace{-1ex}
\hspace{0.3cm} $[500.0, 2000.0)$: \labeltag{warm}
  
\vspace{-1ex}
\hspace{0.3cm} $[2000.0, 5000.0)$: \labeltag{bright}
  
\vspace{-1ex}
\hspace{0.3cm} $[5000.0, 10000.0)$: \labeltag{brilliant}

\vspace{-1ex}
\hspace{0.3cm} $[10000.0, \infty)$: \labeltag{sizzling}

\feat{spectral\_flatness}

\vspace{-1ex}
\hspace{0.3cm} $[0.0, 0.001)$: \labeltag{pure-tone}

\vspace{-1ex}
\hspace{0.3cm} $[0.001, 0.01)$: \labeltag{near-tonal}

\vspace{-1ex}
\hspace{0.3cm} $[0.01, 0.1)$: \labeltag{semi-tonal}
  
\vspace{-1ex}
\hspace{0.3cm} $[0.1, 0.4)$: \labeltag{noise-heavy}
  
\vspace{-1ex}
\hspace{0.3cm} $[0.4, \infty)$: \labeltag{white-noise}

\feat{spectral\_rolloff\_hz}

\vspace{-1ex}
\hspace{0.3cm} $[0.0, 200.0)$: \labeltag{deep-ceiling}

\vspace{-1ex}
\hspace{0.3cm} $[200.0, 1000.0)$: \labeltag{low-ceiling}

\vspace{-1ex}
\hspace{0.3cm} $[1000.0, 5000.0)$: \labeltag{mid-ceiling}
  
\vspace{-1ex}
\hspace{0.3cm} $[5000.0, 12000.0)$: \labeltag{high-ceiling}
  
\vspace{-1ex}
\hspace{0.3cm} $[12000.0, \infty)$: \labeltag{open-ceiling}

\feat{spectral\_flux}

\vspace{-1ex}
\hspace{0.3cm} $[0.0, 0.5)$: \labeltag{frozen}

\vspace{-1ex}
\hspace{0.3cm} $[0.5, 1.5)$: \labeltag{drifting}

\vspace{-1ex}
\hspace{0.3cm} $[1.5, 3.0)$: \labeltag{churning}
  
\vspace{-1ex}
\hspace{0.3cm} $[3.0, 6.0)$: \labeltag{surging}
  
\vspace{-1ex}
\hspace{0.3cm} $[6.0, \infty)$: \labeltag{volatile}

\feat{spectral\_kurtosis}

\vspace{-1ex}
\hspace{0.3cm} $[0.0, 3.0)$: \labeltag{flat-topped}

\vspace{-1ex}
\hspace{0.3cm} $[3.0, 30.0)$: \labeltag{gentle-peak}

\vspace{-1ex}
\hspace{0.3cm} $[30.0, 300.0)$: \labeltag{concentrated}
  
\vspace{-1ex}
\hspace{0.3cm} $[300.0, 3000.0)$: \labeltag{towering}
  
\vspace{-1ex}
\hspace{0.3cm} $[3000.0, \infty)$: \labeltag{needle-point}

\feat{spectral\_entropy}

\vspace{-1ex}
\hspace{0.3cm} $[0.0, 0.15)$: \labeltag{crystalline}

\vspace{-1ex}
\hspace{0.3cm} $[0.15, 0.35)$: \labeltag{ordered}

\vspace{-1ex}
\hspace{0.3cm} $[0.35, 0.6)$: \labeltag{semi-diffuse}
  
\vspace{-1ex}
\hspace{0.3cm} $[0.6, 0.85)$: \labeltag{diffuse}
  
\vspace{-1ex}
\hspace{0.3cm} $[0.85, \infty)$: \labeltag{chaotic}

\feat{spectral\_irregularity}

\vspace{-1ex}
\hspace{0.3cm} $[0.0, 0.02)$: \labeltag{glass-smooth}

\vspace{-1ex}
\hspace{0.3cm} $[0.02, 0.1)$: \labeltag{even-contour}

\vspace{-1ex}
\hspace{0.3cm} $[0.1, 0.3)$: \labeltag{rippled}
  
\vspace{-1ex}
\hspace{0.3cm} $[0.3, 0.55)$: \labeltag{serrated}
  
\vspace{-1ex}
\hspace{0.3cm} $[0.55, \infty)$: \labeltag{comb-like}

\feat{harmonic\_noise\_ratio\_db}

\vspace{-1ex}
\hspace{0.3cm} $\mathrm{NaN}$: \labeltag{unpitched}

\vspace{-1ex}
\hspace{0.3cm} $[-\infty, -3.0)$: \labeltag{noise-engulfed}

\vspace{-1ex}
\hspace{0.3cm} $[-3.0, 3.0)$: \labeltag{murky}

\vspace{-1ex}
\hspace{0.3cm} $[3.0, 8.0)$: \labeltag{hazy}
  
\vspace{-1ex}
\hspace{0.3cm} $[8.0, 14.0)$: \labeltag{limpid}
  
\vspace{-1ex}
\hspace{0.3cm} $[14.0, \infty)$: \labeltag{pristine}

\feat{inharmonicity}

\vspace{-1ex}
\hspace{0.3cm} $\mathrm{NaN}$: \labeltag{unpitched}

\vspace{-1ex}
\hspace{0.3cm} $[0.0, 0.001)$: \labeltag{locked}

\vspace{-1ex}
\hspace{0.3cm} $[0.001, 0.005)$: \labeltag{finely-tuned}

\vspace{-1ex}
\hspace{0.3cm} $[0.005, 0.02)$: \labeltag{slightly-detuned}
  
\vspace{-1ex}
\hspace{0.3cm} $[0.02, 0.1)$: \labeltag{stretched}
  
\vspace{-1ex}
\hspace{0.3cm} $[0.1, \infty)$: \labeltag{warped}

\feat{tristimulus\_1}

\vspace{-1ex}
\hspace{0.3cm} $\mathrm{NaN}$: \labeltag{unpitched}

\vspace{-1ex}
\hspace{0.3cm} $[0.0, 0.3)$: \labeltag{recessed-fundamental}

\vspace{-1ex}
\hspace{0.3cm} $[0.3, 0.6)$: \labeltag{balanced-fundamental}

\vspace{-1ex}
\hspace{0.3cm} $[0.6, 0.85)$: \labeltag{dominant-fundamental}
  
\vspace{-1ex}
\hspace{0.3cm} $[0.85, \infty)$: \labeltag{solo-fundamental}

\feat{tristimulus\_2}

\vspace{-1ex}
\hspace{0.3cm} $\mathrm{NaN}$: \labeltag{unpitched}

\vspace{-1ex}
\hspace{0.3cm} $[0.0, 0.1)$: \labeltag{hollow-body}

\vspace{-1ex}
\hspace{0.3cm} $[0.1, 0.25)$: \labeltag{thin-body}

\vspace{-1ex}
\hspace{0.3cm} $[0.25, 0.4)$: \labeltag{present-body}
  
\vspace{-1ex}
\hspace{0.3cm} $[0.4, \infty)$: \labeltag{lush-body}

\feat{tristimulus\_3}

\vspace{-1ex}
\hspace{0.3cm} $\mathrm{NaN}$: \labeltag{unpitched}

\vspace{-1ex}
\hspace{0.3cm} $[0.0, 0.05)$: \labeltag{bare-upper}

\vspace{-1ex}
\hspace{0.3cm} $[0.05, 0.15)$: \labeltag{sparse-overtones}

\vspace{-1ex}
\hspace{0.3cm} $[0.15, 0.3)$: \labeltag{moderate-overtones}
  
\vspace{-1ex}
\hspace{0.3cm} $[0.3, \infty)$: \labeltag{rich-overtones}

\feat{odd\_even\_harmonic\_ratio}

\vspace{-1ex}
\hspace{0.3cm} $\mathrm{NaN}$: \labeltag{unpitched}

\vspace{-1ex}
\hspace{0.3cm} $[0.0, 0.5)$: \labeltag{even-biased}

\vspace{-1ex}
\hspace{0.3cm} $[0.5, 1.5)$: \labeltag{balanced-parity}

\vspace{-1ex}
\hspace{0.3cm} $[1.5, 5.0)$: \labeltag{odd-leaning}
  
\vspace{-1ex}
\hspace{0.3cm} $[5.0, 50.0)$: \labeltag{odd-heavy}
  
\vspace{-1ex}
\hspace{0.3cm} $[50.0, \infty)$: \labeltag{fundamentals-only}

\feat{sharpness\_acum}

\vspace{-1ex}
\hspace{0.3cm} $[0.0, 1.3)$: \labeltag{dull}

\vspace{-1ex}
\hspace{0.3cm} $[1.3, 2.0)$: \labeltag{mellow}

\vspace{-1ex}
\hspace{0.3cm} $[2.0, 3.0)$: \labeltag{keen}

\vspace{-1ex}
\hspace{0.3cm} $[3.0, 4.5)$: \labeltag{cutting}

\vspace{-1ex}
\hspace{0.3cm} $[4.5, \infty)$: \labeltag{piercing}

\feat{roughness}

\vspace{-1ex}
\hspace{0.3cm} $[0.0, 0.01)$: \labeltag{silky}

\vspace{-1ex}
\hspace{0.3cm} $[0.01, 0.15)$: \labeltag{sleek}

\vspace{-1ex}
\hspace{0.3cm} $[0.15, 0.4)$: \labeltag{textured}

\vspace{-1ex}
\hspace{0.3cm} $[0.4, 0.7)$: \labeltag{gritty}

\vspace{-1ex}
\hspace{0.3cm} $[0.7, \infty)$: \labeltag{abrasive}

\subsubsection{Bark features}

\feat{bark\_levels} (shared by all \feat{bark\_band\_*} features):

\hspace{0.3cm} $[0.0, 0.01)$: \labeltag{silent}

\vspace{-1ex}
\hspace{0.3cm} $[0.01, 2.0)$: \labeltag{trace}

\vspace{-1ex}
\hspace{0.3cm} $[2.0, 5.0)$: \labeltag{faint}
  
\vspace{-1ex}
\hspace{0.3cm} $[5.0, 8.0)$: \labeltag{present}
  
\vspace{-1ex}
\hspace{0.3cm} $[8.0, 11.0)$: \labeltag{strong}

\vspace{-1ex}
\hspace{0.3cm} $[11.0, 15.0)$: \labeltag{dominant}

\vspace{-1ex}
\hspace{0.3cm} $[15.0, \infty)$: \labeltag{overwhelming}

\feat{bark\_bands} (band index $\to$ keyword):

\vspace{-1ex}
\hspace{0.3cm} $\mathrm{band}\ 1$: \labeltag{rumble}

\vspace{-1ex}
\hspace{0.3cm} $\mathrm{band}\ 2$: \labeltag{thump}

\vspace{-1ex}
\hspace{0.3cm} $\mathrm{band}\ 3$: \labeltag{boom}

\vspace{-1ex}
\hspace{0.3cm} $\mathrm{band}\ 4$: \labeltag{boxiness}

\vspace{-1ex}
\hspace{0.3cm} $\mathrm{band}\ 5$: \labeltag{honk}

\vspace{-1ex}
\hspace{0.3cm} $\mathrm{band}\ 6$: \labeltag{quack}

\vspace{-1ex}
\hspace{0.3cm} $\mathrm{band}\ 7$: \labeltag{clang}

\vspace{-1ex}
\hspace{0.3cm} $\mathrm{band}\ 8$: \labeltag{punch}

\vspace{-1ex}
\hspace{0.3cm} $\mathrm{band}\ 9$: \labeltag{bite}

\vspace{-1ex}
\hspace{0.3cm} $\mathrm{band}\ 10$: \labeltag{twang}

\vspace{-1ex}
\hspace{0.3cm} $\mathrm{band}\ 11$: \labeltag{ring}

\vspace{-1ex}
\hspace{0.3cm} $\mathrm{band}\ 12$: \labeltag{tang}

\vspace{-1ex}
\hspace{0.3cm} $\mathrm{band}\ 13$: \labeltag{edge}

\vspace{-1ex}
\hspace{0.3cm} $\mathrm{band}\ 14$: \labeltag{chime}

\vspace{-1ex}
\hspace{0.3cm} $\mathrm{band}\ 15$: \labeltag{zing}

\vspace{-1ex}
\hspace{0.3cm} $\mathrm{band}\ 16$: \labeltag{crackle}

\vspace{-1ex}
\hspace{0.3cm} $\mathrm{band}\ 17$: \labeltag{sibilance}

\vspace{-1ex}
\hspace{0.3cm} $\mathrm{band}\ 18$: \labeltag{fizz}

\vspace{-1ex}
\hspace{0.3cm} $\mathrm{band}\ 19$: \labeltag{sheen}

\vspace{-1ex}
\hspace{0.3cm} $\mathrm{band}\ 20$: \labeltag{sparkle}

\vspace{-1ex}
\hspace{0.3cm} $\mathrm{band}\ 21$: \labeltag{glint}

\vspace{-1ex}
\hspace{0.3cm} $\mathrm{band}\ 22$: \labeltag{air}

\vspace{-1ex}
\hspace{0.3cm} $\mathrm{band}\ 23$: \labeltag{vapor}

\vspace{-1ex}
\hspace{0.3cm} $\mathrm{band}\ 24$: \labeltag{ether}

Bark band labels are composed as \{\labeltag{level}\} \{\labeltag{keyword}\}. For example, for bark band $6$ and level $[15.0,\infty)$, we get an \labeltag{overwhelming quack}.

\subsubsection{Fundamental frequency features (\feat{f0\_hz})}

Except for NaN (mapped to \labeltag{unpitched}), these labels are generated procedurally. 

There are $8 \times 12 \times 3 = 288$ finite bins, covering $[0,5120)$ Hz in $36$ equal-ratio bins per octave. Let $r \in \{0,\dots,7\}$ be the register index, $n \in \{0,\dots,11\}$ the chromatic index, and $m \in \{0,1,2\}$ the microstep index.
Define the global bin index
\[
k = 36r + 3n + m.
\]
The corresponding interval is
\[
I_k =
\begin{cases}
[0.0,\ 20 \cdot 2^{1/36}), & k = 0,\\[0.5ex]
[20 \cdot 2^{k/36},\ 20 \cdot 2^{(k+1)/36}), & k = 1,\dots,287.
\end{cases}
\]
Its lexical label is composed as
\[
\{\labeltag{register}(r)\}\ \{\labeltag{chromatic}(n)\}\ \{\labeltag{micro}(m)\}.
\]

\feat{f0\_registers} (register index $\to$ word):

\vspace{-1ex}
\hspace{0.3cm} $0$: \labeltag{sub}

\vspace{-1ex}
\hspace{0.3cm} $1$: \labeltag{cellar}

\vspace{-1ex}
\hspace{0.3cm} $2$: \labeltag{chest}

\vspace{-1ex}
\hspace{0.3cm} $3$: \labeltag{middle}

\vspace{-1ex}
\hspace{0.3cm} $4$: \labeltag{lumen}

\vspace{-1ex}
\hspace{0.3cm} $5$: \labeltag{aloft}

\vspace{-1ex}
\hspace{0.3cm} $6$: \labeltag{crystal}

\vspace{-1ex}
\hspace{0.3cm} $7$: \labeltag{stratos}

\feat{f0\_chromatic} (chromatic index $\to$ word):

\vspace{-1ex}
\hspace{0.3cm} $0$: \labeltag{do}

\vspace{-1ex}
\hspace{0.3cm} $1$: \labeltag{di}

\vspace{-1ex}
\hspace{0.3cm} $2$: \labeltag{re}

\vspace{-1ex}
\hspace{0.3cm} $3$: \labeltag{ri}

\vspace{-1ex}
\hspace{0.3cm} $4$: \labeltag{mi}

\vspace{-1ex}
\hspace{0.3cm} $5$: \labeltag{fa}

\vspace{-1ex}
\hspace{0.3cm} $6$: \labeltag{fi}

\vspace{-1ex}
\hspace{0.3cm} $7$: \labeltag{sol}

\vspace{-1ex}
\hspace{0.3cm} $8$: \labeltag{si}

\vspace{-1ex}
\hspace{0.3cm} $9$: \labeltag{la}

\vspace{-1ex}
\hspace{0.3cm} $10$: \labeltag{li}

\vspace{-1ex}
\hspace{0.3cm} $11$: \labeltag{ti}

\feat{f0\_micro} (microstep index $\to$ word):

\vspace{-1ex}
\hspace{0.3cm} $0$: \labeltag{shadow}

\vspace{-1ex}
\hspace{0.3cm} $1$: \labeltag{heart}

\vspace{-1ex}
\hspace{0.3cm} $2$: \labeltag{crown}

For example:
\begin{itemize}
    \item $(r,n,m)=(0,0,0)$ gives \labeltag{sub do shadow};
    \item $(r,n,m)=(1,0,0)$ gives \labeltag{cellar do shadow};
    \item $(r,n,m)=(4,7,2)$ gives \labeltag{lumen sol crown}.
\end{itemize}

\section{Decoder implementation details}
\label{app:decoder-details}
We report additional implementation details that are useful for reproduction but are not needed to understand the main decoding pipeline.

\emph{Code and data for full reproducibility will be made publicly available upon acceptance.}

\paragraph{Seed.}
The seed $\sigma$ is defined from the recovered lexical code $\ell$ as $\sigma=\mathrm{hash}(\ell)$. This way, paraphrasing at the sentence level does not change the sound as long as the recovered code is the same. 

In our implementation, the seed is computed once at the start of decoding by hashing the canonical token sequence with \texttt{SHA-256} and taking the first $32$ bits. 
It is computed before refinement, and the same seed is reused for every synthesis call during refinement and for the final render. That makes the stochastic renderer reproducible and keeps the optimization objective stable instead of changing randomly across evaluations.

\paragraph{Duration.}
The decoded attack time is obtained from the log-attack representative. Undefined attack falls back to 1 ms, while decay time falls back to 0.5 s. The renderer uses
\[
    \mathrm{duration}=\mathrm{attack}+4\cdot \mathrm{decay},
\]
clamped to $[0.05,5.0]$ seconds. If a target sample length is supplied, the attack and decay constants are rescaled so that the output has exactly that length while preserving their relative shape.

\paragraph{Renderer controls.}
Our current implementation uses a 15-dimensional control vector $c$ with the following parameters:
\begin{itemize}
    \item \textbf{source gains ($4\times$):} harmonic, modal, noise, transient;
    \item \textbf{temporal scales ($3\times$):} transient decay, noise decay, body decay;
    \item \textbf{resonance / texture controls ($2\times$):} modal density, roughness;
    \item \textbf{spectral-shaping controls ($6\times$):} body pivot, transient brightness, spectral tilt, low emphasis, high emphasis, spectral spread shape.
\end{itemize}
These controls are initialized deterministically from the decoded features. 

We emphasize that these are renderer controls meant to simplify decoding, \emph{not} additional acoustic descriptors. They admit different implementations and are not required to follow our exact recipe. 

When instructing a coding agent, a possible prompt could be: 
\begin{center}
\emph{"Implement a hybrid renderer with harmonic, modal, body-noise, and transient-noise layers, then expose a small set of monotonic macro-controls that scale source mixture, decay times, resonance density, roughness, and broad spectral shaping."}
\end{center}
More details for improving stability and predictability are given below (exact formulas can differ across implementations):
\begin{itemize}
    \item \textbf{Harmonic gain:} multiplies the pitched harmonic layer before mixing. Increasing it makes the sound more tonal, periodic, and pitch-dominant.
    \item \textbf{Modal gain:} multiplies the bank of damped resonant modes. Increasing it strengthens body-like resonance and pitched or quasi-pitched ringing that is not strictly harmonic.
    \item \textbf{Noise gain:} multiplies the body-noise layer. Increasing it raises broadband noisy energy in the sustained part of the sound.
    \item \textbf{Transient gain:} multiplies the transient-noise layer. Increasing it makes the onset sharper, noisier, and more attack-heavy.

    \item \textbf{Transient decay scale:} scales the decay time of the transient envelope. Larger values produce longer, more extended attacks; smaller values produce shorter and more percussive attacks.
    \item \textbf{Noise decay scale:} scales the decay of the broadband noise component. Larger values keep noisy energy present for longer.
    \item \textbf{Body decay scale:} scales the decay of the resonant/body portion of the sound. Larger values produce longer ringing or sustain.

    \item \textbf{Modal density:} controls how many modal resonances are active, or how densely they fill the frequency axis. Increasing it makes the resonant layer thicker and more diffuse.
    \item \textbf{Roughness:} controls local detuning, beating, jitter, or nearby companion resonances. Increasing it makes the sound harsher, buzzier, or more beating-rich.

    \item \textbf{Body pivot:} sets the pivot frequency around which broad spectral shaping is applied. Intuitively, it determines the frequency region relative to which the body is made darker or brighter.
    \item \textbf{Transient brightness:} increases high-frequency emphasis in the transient layer. Larger values yield a crisper or splashier onset.
    \item \textbf{Spectral tilt:} applies a broadband slope to the spectrum, typically in the log-frequency domain. Increasing it shifts energy toward high frequencies; decreasing it shifts energy toward low frequencies.
    \item \textbf{Low emphasis:} applies broad low-frequency boost or attenuation, similar to a coarse low-shelf control.
    \item \textbf{High emphasis:} applies broad high-frequency boost or attenuation, similar to a coarse high-shelf control.
    \item \textbf{Spectral spread shape:} controls how concentrated or diffuse energy is around the main spectral mass. Increasing it broadens the spectrum; decreasing it makes the spectrum more compact.
\end{itemize}

A useful way to think about these controls is that they act on mechanisms, not directly on features. For example, \emph{spectral tilt}, \emph{high emphasis}, and \emph{transient brightness} all influence measured descriptors such as centroid, rolloff, flatness, and sharpness, but they do so indirectly through broad spectral shaping. Likewise, \emph{modal density}, \emph{roughness}, and the various gains can influence several measured features at once. This is intentional: the controls are low-dimensional steering variables for the renderer, while the transmitted lexical features remain the external acoustic targets.

A minimal implementation can realize these controls by mixing four source layers (harmonic, modal, body noise, transient noise), applying separate attack/decay envelopes, then applying broad post-mix spectral shaping and RMS normalization.

\paragraph{Hybrid source model.}
The harmonic layer is an additive sine bank with up to 24 partials. Partial frequencies follow
\[
    f_k = k f_0\sqrt{1+\beta k^2},
\]
where $\beta$ is the decoded inharmonicity. Partial amplitudes are allocated from the decoded tristimulus coefficients and adjusted by the odd/even harmonic ratio. The modal layer places damped sinusoids near Bark-band center frequencies. The noise layers are deterministic white-noise bursts seeded from the lexical code and shaped with Bark-domain equalization. All layers are shaped by linear-attack/exponential-decay envelopes and mixed according to the current renderer controls.




\end{document}